\documentclass[10pt,twocolumn,letterpaper]{article}

\usepackage{iccv}
\usepackage{times}
\usepackage{epsfig}
\usepackage{graphicx}
\usepackage{amsmath}
\usepackage{amssymb}
\usepackage{multirow}
\usepackage{graphicx}
\usepackage{caption}
\usepackage{subfigure}
\usepackage{array}
\usepackage{verbatim}
\usepackage{float}
\usepackage[T1]{fontenc}
\usepackage[utf8]{inputenc}
\usepackage{authblk}

\usepackage[breaklinks=true,bookmarks=false]{hyperref}

\iccvfinalcopy % *** Uncomment this line for the final submission

\def\iccvPaperID{5288} % *** Enter the ICCV Paper ID here

\ificcvfinal\fi
\begin{document}
	\bibliographystyle{unsrt}
	%%%%%%%%% TITLE
	\title{Knowledge Squeezed Adversarial Network Compression}

	\author[2]{Changyong Shu}
	\author[2]{Peng Li}
	\author[1]{Yuan Xie \thanks{Corresponding author: xieyuan8589@foxmail.com}}
	\author[3]{Yanyun Qu}
	\author[4]{Longquan Dai}
	\author[5]{Lizhuang Ma}

	\affil[1]{East China Normal University, xieyuan8589@foxmail.com}
	\affil[2]{Nanjing Institute of Advanced Artificial Intelligence, \{changyong.shu, peng03.li\}@horizon.ai}
	\affil[3]{XMU, yyqu@xmu.edu.cn}
	\affil[4]{Nanjing University of Science and Technology, lqdai@foxmail.com}
	\affil[5]{Shanghai Jiao Tong University, ma-lz@cs.sjtu.edu.cn}
%	\thanks{\, xieyuan8589@foxmail.com, yyqu@xmu.edu.cn, lqdai@foxmail.com, ma-lz@cs.sjtu.edu.cn}

	\maketitle
	%\thispagestyle{empty}

	%%%%%%%%% ABSTRACT
	\begin{abstract}
	Deep network compression has been achieved notable progress via knowledge distillation, where a teacher-student learning manner is adopted by using predetermined loss. Recently, more focuses have been transferred to employ the adversarial training to minimize the discrepancy between distributions of output from two networks. However, they always emphasize on result-oriented learning while neglecting the scheme of process-oriented learning, leading to the loss of rich information contained in the whole network pipeline. Inspired by the assumption that, the small network can not perfectly mimic a large one due to the huge gap of network scale, we propose a knowledge transfer method, involving effective intermediate supervision, under the adversarial training framework to learn the student network. To achieve powerful but highly compact intermediate information representation, the squeezed knowledge is realized by task-driven attention mechanism. Then, the transferred knowledge from teacher network could accommodate the size of student network. As a result, the proposed method integrates merits from both process-oriented and result-oriented learning. Extensive experimental results on three typical benchmark datasets, {\it i.e.,} CIFAR-10, CIFAR-100, and ImageNet, demonstrate that our method achieves highly superior performances against other state-of-the-art methods.
	\end{abstract}
	%%%%%%%%% BODY TEXT
	\section{Introduction}
	Deep neural networks (DNNs) greatly enhance the development of artificial intelligence via dominant performance in diverse perception missions. Whereas the computing resource-consuming problem omnipresent in modern DNNs restricts the relative implementation on embedded systems, which further expedites the development of network compression. The network compression can accelerate neural networks for real-time applications on edge devices via the following representative aspects: low-rank decomposition \cite{tai2015convolutional,denton2014exploiting}, network pruning \cite{Song2015Deep,Hao2016Pruning,han2015deep}, quantization \cite{courbariaux2016binarized,Zhou2017Incremental}, knowledge distillation (KD) \cite{hinton2015distilling,Luo2016Face}, and compact network design \cite{Iandola2016SqueezeNet,howard2017mobilenets,zhang2018shufflenet}.
	
	\begin{figure}[t]
		\begin{center}
			\includegraphics[width=0.9\linewidth]{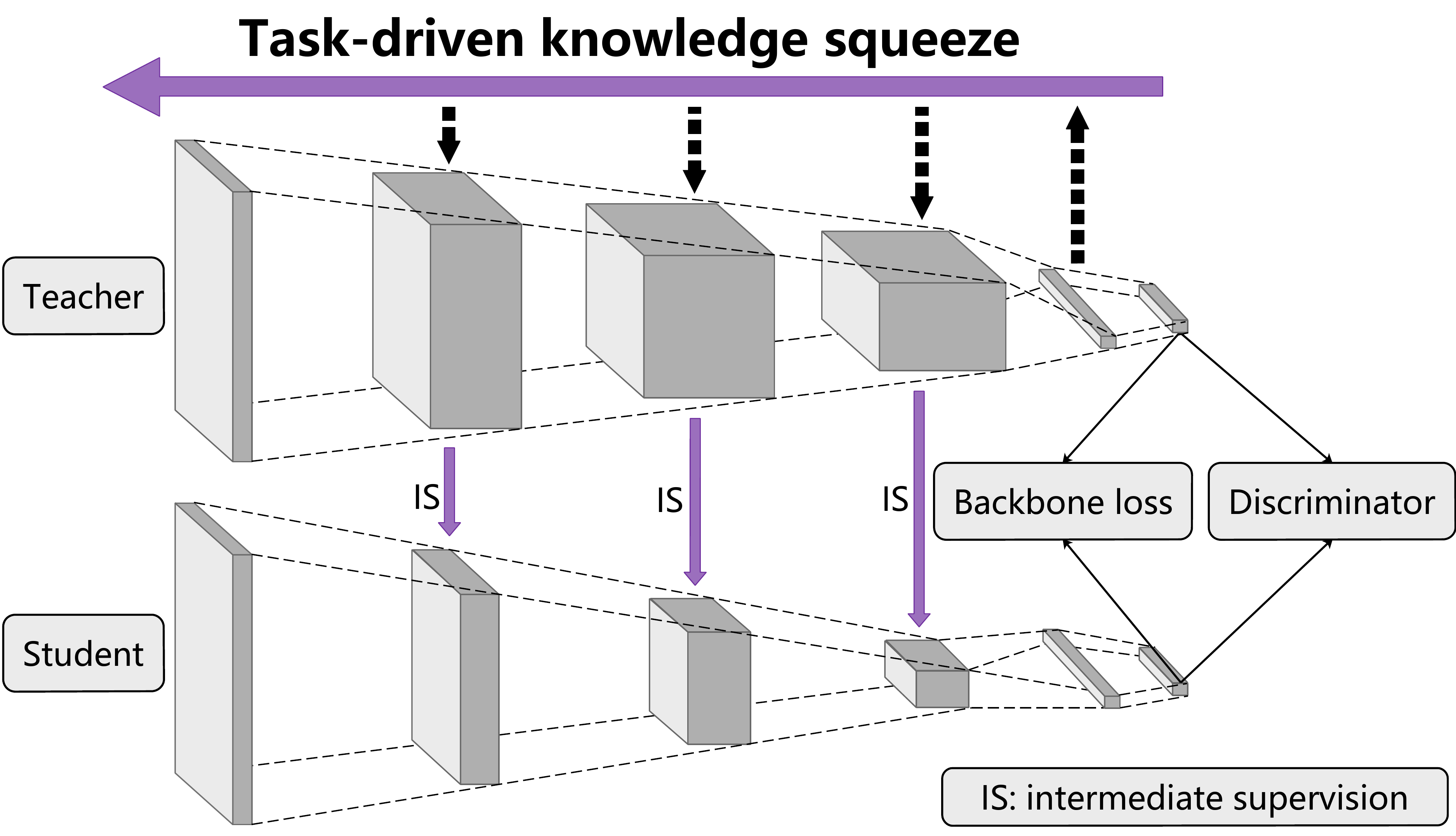}
		\end{center}
		\caption{The overview of proposed Knowledge Squeezed Adversarial Network Compression (KSANC) via intermediate supervision.}
		\label{fig:long}
		\label{fig:arch}
	\end{figure}
	
	Among the above categories, KD is somewhat different due to the utilization of information from the pre-trained teacher network. Hinton \etal \cite{hinton2015distilling} forced the output of student network to match the soft targets produced by a teacher network via KL divergence. Through promising results have been achieved, it is difficult to determine which metric is optimal to measure the information inherited from teacher \cite{inproceedings}. As the loss learned by adversarial training has advantage over the predetermined loss in the student-teacher strategy empirically, Belagiannis \etal \cite{ANC} and Xu \etal \cite{tstn} proposed the GAN-based distillation approach by introducing the discriminator in GAN to match the output distribution between teacher and student.  

	Although the high level distilled knowledge has been explored in both of the approaches mentioned above with relatively acceptable performance, considering the case of teaching of human being, {\it does it the best methodology for teacher to teach the student by only giving the final result of a problem?}
	
	\textbf{Motivations: } 1) To inherit the information from teacher network, the aforementioned methods \cite{ANC,tstn,inproceedings} usually focus on result-oriented learning, which utilizes the adversarial training to minimize the discrepancy between distributions of extracted feature ({\it{i.e.}}, the input vector of the softmax layer) or the unscaled log-probability values ({\it{i.e.}}, logits) from teacher and student networks. While reasonably effective, rich information encoded in the intermediate layers of teacher might be ignored. Therefore, we aim to add the intermediate supervision to conduct knowledge transfer, such that both process-oriented and result-oriented learning can complement each other.
	
	2) Even thought a lot of redundancy existed in large network, {\it we should acknowledge that the small network can not mimic a large one perfectly}, due to the significant difference in the number of layers. For example, ResNet-164 and ResNet-20 achieve error rate of $27.76\%$ and $33.36\%$ on CIFAR-100 (see Table \ref{tab:sofa-cifar}), respectively, which demonstrates that there is a large gap of representation capability between these two architectures. To reduce this gap, besides result learning, we have to squeeze the knowledge contained in the whole pipeline of the teacher network into a compact form to accommodate the size of student network. With the help of the task-driven attention mechanism, the so called knowledge squeeze can be realized in an elegant manner. Moreover, by incorporating the intermediate supervision, the squeezed knowledge can be effectively injected into the student network.
	
	In this paper, we present the Knowledge Squeezed Adversarial Network Compression (KSANC) via intermediate supervision, as shown in Figure \ref{fig:arch}. The major contributions can be summarized as follows: (i) A novel knowledge transfer method, which involves the effective intermediate supervision, is proposed based on the adversarial training framework to implement the deep network compression in a teacher-student manner. (ii) To achieve powerful intermediate representation, we introduce the task-driven attention mechanism to squeeze the knowledge contained in teacher network toward the lower layers so as to accommodate the size of student network. (iii) We conduct the extensive evaluation of our method on several representative datasets, where the experimental results demonstrate that our method achieves highly competitive performance compared with other knowledge transfer approaches, while maintaining smaller accuracy drop.
			
	%------------------------------------------------------------------------
	\section{Related Work}
	
	Our work mostly relates to the model compression, knowledge distillation, and attention transfer, we will describe each of them respectively.
	
	%-------------------------------------------------------------------------
	\subsection{Network Compression}
	We briefly review the following five kinds of approach for deep network compression. (1) Low rank decomposition: In this case, the main idea is to construct low rank basis of filters to effectively reduce the weight tensor. For instance, Jaderberg \etal \cite{jaderberg2014speeding} proposed an agnostic approach to have rank-1 filters in the spatial domain. Related approaches have also be explored by the same principle of finding a low-rank approximation for the convolutional layers \cite{rigamonti2013learning,denton2014exploiting,lebedev2014speeding,yang2015deep}. (2) Network pruning: Removing network connections not only reduces the model size but also prevents over-fitting. Parameter sharing has also contributed to reduce the network parameters with repetitive patterns \cite{belagiannis2017recurrent,schmidhuber1992learning}. (3) Quantization:  Quantization reduces the size of memory requirement and accelerates the inference by using weights with discrete values \cite{soudry2014expectation,wu2016quantized,rastegari2016xnor,courbariaux2015binaryconnect,li2016ternary}. (4) Compact network design: Many researchers resort to derive more efficient network architectures, such as ResNets \cite{he2016deep}, SqueezeNet \cite{Iandola2016SqueezeNet} and ShuffleNet \cite{zhang2018shufflenet}, to shrink the number of the parameters while maintaining the performance. (5) Distillation: The proposed method belongs to this category, which will be discussed in detail in the next subsection.
	
	\begin{figure*}
		\begin{center}
			%\fbox{\rule{0pt}{2in} \rule{.9\linewidth}{0pt}}
			%\includegraphics[width=0.8\linewidth]{fig_2.eps}
			\includegraphics[width=1.0\linewidth]{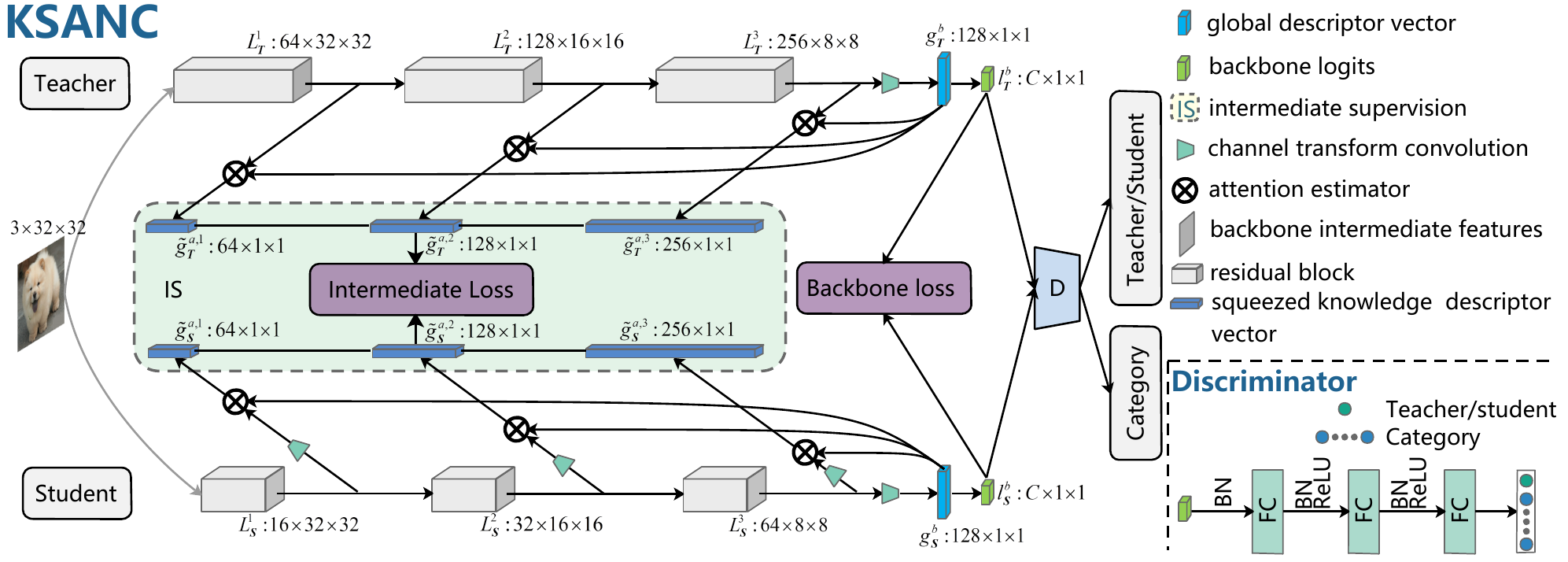}
		\end{center}
		\caption{The architecture of our proposed KSANC. The sub-figure in the upleft of the dotted line is the paradigm of our teacher-student strategy. The sub-figure in the bottom right corner of the dotted line is the structure of the discriminator. This is an example where teacher is ResNet-164 and student is ResNet-20, and $C$ is the category number.}
		\label{fig:overview}
	\end{figure*}

	%-------------------------------------------------------------------------
	\subsection{Knowledge Distillation}
	Knowledge distillation \cite{ba2014deep,li2014learning,polino2018model} is used to transfer knowledge from teacher network to student network by the output before the softmax function (logits) or after it (soft targets), which has been popularized by Hinton \etal \cite{hinton2015distilling}. Various derivatives have been implemented in different tasks: object detection \cite{Liu2018KTAN}, semantic segmentation \cite{ros2016training}, and metric learning \cite{chen2018darkrank}, etc. As it is hard for student network with small capacity to mimic the outputs of teacher network, several researches \cite{ANC,tstn} focused on using adversarial networks to replace the manually designed metric such as ${\it{L}}1$/${\it{L}}2$ loss or KL divergence.	
		
	While our approach is complementary to the above result-oriented learning methods by introducing the process-oriented learning strategy, which pays more attention on transferring intermediate knowledge \cite{romero2014fitnets,agift,attmap,Zhou2017Rocket} instead of final result from teacher to student.
	%-------------------------------------------------------------------------
	\subsection{Attention Transfer}
	Early work on attention based tracking \cite{larochelle2010learning,denil2012learning} was motivated by human attention mechanism theories \cite{rensink2000dynamic}, and was accomplished via Restricted Bolzmann Machines. It was exploited in computer-vision-related tasks such as image captioning \cite{xu2015show}, visual question answering \cite{yang2016stacked}, as well as in weakly-supervised object localization \cite{oquab2015object} and classification \cite{mnih2014recurrent}. In transfer learning,  attention transfer can facilitate the fast optimization and improve the performance of a small student network via the attention map \cite{attmap} or the flow of solution procedure (FSP) matrix \cite{agift}. Attention transfer is also introduced in machine reading comprehension \cite{hu2018attention} to transfer teacher attentive information.
	
	The above transfer methods emphasize on how to represent the intermediate information more effectively, while neglecting to construct a compact form of knowledge for transfer, which is more important in compression since we have acknowledged that the capability of the small network is far below the deep one. Different from them, we propose to squeeze the intermediate knowledge by using a task-driven attention \cite{Jetley2018Learn}, such that the compact knowledge from teacher network could accommodate the size of student network.

	%------------------------------------------------------------------------
	\section{The Proposed Method}
	In this section, we present how to achieve the knowledge transfer between teacher and student networks via the framework of knowledge squeezed adversarial training. We start by giving the architecture overview, and then elaborate each term in loss function and depict the optimization method in subsection \ref{loss-KSANC} and \ref{opt-KSANC}, respectively.
	%-------------------------------------------------------------------------
	\subsection{The Architecture of KSANC}\label{arch-KSANC}
	As illustrated in Figure \ref{fig:overview}, our method consists of the teacher, student and discriminator networks. We denote the teacher network (require pre-train) as $\boldsymbol{T}$, and the student network as $\boldsymbol{S}$. Both the teacher and student network are built by a backbone-subnetwork (\eg, VGG, ResNet) $\text{\it{Net}}_{b}$ and an attention-subnetwork $\text{\it{Net}}_{a}$. The $\text{\it{Net}}_{b}$ is a standard CNN pipelines with $N$ ($N=3$ in Figure \ref{fig:overview}) replicated blocks, where we can obtain $N$ corresponding intermediate features $L_{\boldsymbol{T/S}} = \{L_{\boldsymbol{T/S}}^{1}, L_{\boldsymbol{T/S}}^{2},\ldots, L_{\boldsymbol{T/S}}^{N}\}$.
	
\begin{figure}[t]
		\begin{center}
			\includegraphics[width=1.0\linewidth]{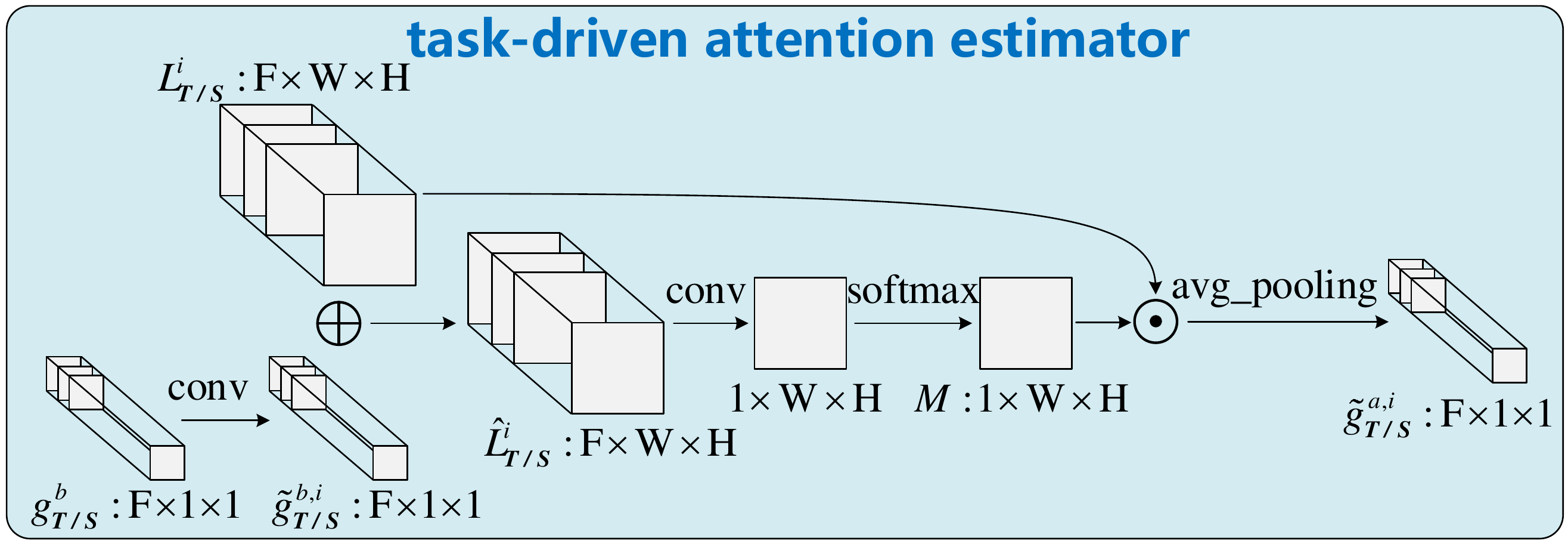}
		\end{center}
		\caption{The structure of designed task-driven attention estimator for $i$-th block, where the backbone intermediate feature $L_{\boldsymbol{T/S}}^{i}$ and the global descriptor $g^{b}_{\boldsymbol{T/S}}$ are taken as input to produce the squeezed knowledge descriptor $\widetilde{g}^{a,i}_{\boldsymbol{T/S}}$.}
		\label{fig:attention-estimator}
	\end{figure}

	In attention-subnetwork, the attention estimator (the fifth line in the legend of Figure \ref{fig:overview}) can produce the squeezed knowledge descriptor $\widetilde{g}^{a,i}_{\boldsymbol{T/S}}$ ($i=1,\ldots,N$, the blue vectors in the dashed box in Figure \ref{fig:overview}) corresponding to the $i$-th block by taking the backbone intermediate feature $L_{\boldsymbol{T/S}}^{i}$ and the global descriptor $g^{b}_{\boldsymbol{T/S}}$ ({\it i.e.}, the input vector of the softmax layer) as the input. The detailed structure of the task-driven attention estimator is presented in Figure \ref{fig:attention-estimator}. As the channel number of $g^{b}_{\boldsymbol{T/S}}$ might differ from that of $L_{\boldsymbol{T/S}}^{i}$, the channel alignment convolution is introduced to achieve the feature vector $\hat{g}^{b,i}_{\boldsymbol{T/S}}$. Then, the element-wise sum between $\hat{g}^{b,i}_{\boldsymbol{T/S}}$ and $L_{\boldsymbol{T/S}}^{i}$ is conducted along the dimensionality of channel. Finally, the squeezed knowledge descriptor $\widetilde{g}^{a,i}_{\boldsymbol{T/S}}$ for the $i$-th block can be gained by the following equation:
	\begin{equation}
	M = \text{softmax}(\widehat{W} * \hat{L}_{\boldsymbol{T/S}}^{i})
	\end{equation}
	\begin{equation}
	\widetilde{g}^{a,i}_{\boldsymbol{T/S}} = \text{average\_pooling}(M \odot \hat{L}_{\boldsymbol{T/S}}^{i})
	\end{equation}
	where the $\widehat{W}$, a $C\times1\times1$ convolution kernel, is used to compute the attention score $M$, and $*$ denotes the convolutional operator, and $\odot$ is the element-wise product.
	
	With the help of attention estimator, the knowledge contained in each backbone block of the teacher network could be converted into a compact form by integrating the task-specific information ({\it i.e.,}  $\widetilde{g}^{a,i}_{\boldsymbol{T/S}}$), such that the so-called knowledge squeeze can be realized.
	
	As for the discriminator, it aims to distinguish where the input vectors ({\it i.e.,} logits, denoted by $l_{\boldsymbol{T/S}}^b$) come from (teacher or student). The structure of the discriminator, as it is shown in the bottom right corner of Figure \ref{fig:overview}, is composed of three sequentially stacked fully-connected layers. The number of nodes in all hidden layers is the dimension of the input. We use backbone logits as the input of discriminator rather than the global descriptor $g_{\boldsymbol{T/S}}^b$ in our experiments, since the discriminator here is considered as a result-oriented guidance to match the output distribution between teacher and student, the logits $l_{\boldsymbol{T/S}}^b$ is more preferable.
	
	%-------------------------------------------------------------------------
	\subsection{Overall Loss Function}\label{loss-KSANC}
	In order to train our network, we define a loss function in Eqn. (\ref{eq:overall_loss}) including three components, {\it i.e.,} the adversarial loss $L_{adv}$, the backbone loss $L_{b}$, and the intermediate loss $L_{is}$:
	\begin{equation}
	L = \lambda_1 L_b + \lambda_2 L_{adv} + \lambda_3 L_{is}
	\label{eq:overall_loss}
	\end{equation}
	where $\lambda_1$, $\lambda_2$, $\lambda_3$ are trade-off factors. During the process of knowledge transfer, the backbone loss $L_{b}$ is utilized to directly match the output of two networks, while the adversarial loss $L_{adv}$ is employed to minimize the discrepancy between distributions of logits from two networks. Both of them are served for result-oriented learning. On the contrary, the intermediate loss $L_{is}$ could facilitate the process-oriented learning via intermediate supervision. Consequently, both process-oriented and result-oriented learning can complement each other. To simplify the model, the weight of each loss are set to be equal, {\it i.e.,} $\lambda_1 = \lambda_2 = \lambda_3 = 1$.
	
	\textbf{Backbone Loss: }Backbone loss is a result-oriented constraint which makes the student mimic the teacher by minimizing the $L2$ loss between teacher and student backbone logits:
	\begin{equation}
	{L_b} = \|{l_{\boldsymbol{S}}^b(x) - l_{\boldsymbol{T}}^b(x)}\|_2^2
	\label{eq:result_oriented_L2}
	\end{equation}
	
	\textbf{Adversarial Loss: }In the proposed model, a GAN based approach is introduced to transfer knowledge from teacher to student. The teacher and student networks are used to convert an image sample $x$ to the logits $l_{\boldsymbol{T}}^b(x)$ and $l_{\boldsymbol{S}}^b(x)$, respectively, where the student is considered as a generator in vanilla GAN. While the discriminator aims to distinguish whether the input logits comes from teacher or student. As our goal is to fool the discriminator in predicting the same output for teacher and student networks, the objective can be written as:
	\begin{equation}
	\begin{split}
	L_{adv}^o = \min\limits_{l_{\boldsymbol{S}}^b} \max\limits_{D} \quad
	& E_{l_{\boldsymbol{T}}^b(x)\sim{p_{{\boldsymbol{T}}}}}[{\log({D({l_{\boldsymbol{T}}^b(x)})})}] + \\
	& E_{l_{\boldsymbol{S}}^b(x)\sim{p_{\boldsymbol{S}}}}[{\log({1 - D({l_{\boldsymbol{S}}^b(x)})})}]
	\end{split}
	\label{eq:ori_adv}
	\end{equation}
	where $p_{{\boldsymbol{T}}}$ and $p_{\boldsymbol{S}}$ correspond to logits distribution of the teacher and student network, respectively.
	
	In order to get more valuable gradient for student, the regularization and category-level supervision are introduced to further improve the discriminator. We utilize three regularizers to prolong the minimax game between the student and discriminator as follows:
	\begin{equation}
	L_{reg} = -\mu\bigg(|\omega_{D}|+\|\omega_{D}\|_2^2- E_{l_{\boldsymbol{S}}^b(x)\sim{p_{\boldsymbol{S}}}}[{\log({D({l_{\boldsymbol{S}}^b(x)})})}]\bigg)
	\label{eq:regularization}
	\end{equation}
	where $\omega_{D}$ is the parameters of discriminator, and $\mu$ controls the contribution of regularizer in optimization, the negative sign denotes that the loss term is updated in the maximization step. The first two terms force the weights of discriminator to grow slowly, the last term is referred to as the adversarial sample regularization, and the above loss terms are original designed in \cite{tstn}. It utilizes the additional student samples (labeled as teacher) to confuse the discriminator such that the capability of discriminator can be restricted to some extent.
	
	Note that the adversarial loss defined in Eqn. (\ref{eq:ori_adv}) only focus on matching the logits on distribution-level, while missing the category information might result in the incorrect association between logits and labels. Consequently, the discriminator is further modified to simultaneously predict the ``teacher/student'' and the class labels. On this occasion, the output of discriminator is a $1+C$ dimensional vector (the first element represents ``teacher/student'', while the remaining denote the category by using one-hot encoding), and the category regularizer for the discriminator can be written as:
	\begin{equation}
	\begin{split}
	L_{adv}^{C} = \min\limits_{l_{\boldsymbol{S}}^b} \max\limits_{D} \quad
	&{E_{l_{\boldsymbol{T}}^b(x)\sim{p_{\boldsymbol{T}}}}}[\log(P(l(x)|C_{\boldsymbol{T}}(x)))]\\
	+ &{E_{l_{\boldsymbol{S}}^b(x)\sim{p_{\boldsymbol{S}}}}}[\log(P(l(x)|C_{\boldsymbol{S}}(x)))]
	\end{split}
	\label{eq:adv_with_cgy}
	\end{equation}
	where $l(x)$ means the label of the sample $x$, $C_{\boldsymbol{T/S}}(x)$ corresponds to $D({l_{\boldsymbol{T/S}}^b(x)})$[1:], where the [\;] denotes the vector slice in python. As the discriminator has to learn with the extended outputs to jointly predict ``teacher/student`` and category, the adversarial learning becomes more stable.
	
	%The $L_{adv}$ in Eq. \ref{eq:adv_with_cgy} solve the problem by maximizing the log-likelihood between true label $l$ and the predicted probability $C_{\boldsymbol{T/S}}(x)$. So the teacher and student output with the same input image are more possibly to belong to the same category.
	
	%Moreover, the original GAN in Eq. \ref{eq:ori_adv} only considers the distributed consensus of the high-level statistics of teacher and student backbone logits, the single example alignment is missing, which means that the teacher and student backbone logits $l_{\boldsymbol{T/S}}^b(x)$ for the same image $x$ may differ in class.
	
	To sum up, the final loss for adversarial training can be formulated as:
	\begin{equation}\label{eq:final-adv-loss}
	L_{adv} = L_{adv}^{o} + L_{reg} + L_{adv}^{C}
	\end{equation}
	
	\textbf{Intermediate Loss: }The loss of intermediate supervision is an effective term to inject the squeezed knowledge, {\it i.e.,} $\widetilde{g}^{a,i}_{\boldsymbol{T}}, (i=1,\ldots,N)$, into the student network. It can be given by the $L2$ distance:
	\begin{equation}
	{L_{is}} = \|{\widetilde{g}_{\boldsymbol{S}}^a(x) - \widetilde{g}_{\boldsymbol{T}}^a(x)}\|_2^2
	\end{equation}
	where $\widetilde{g}_{\boldsymbol{T/S}}^a(x)$ is the concatenation of $\widetilde{g}^{a,i}_{\boldsymbol{T/S}}(i=1,...,N)$. It is noteworthy that other loss function, such as $L1$ loss and cross-entropy loss, can also be applied. Besides, we find that $L2$ loss outperforms others empirically in our experiments.
	
	%-------------------------------------------------------------------------
	\subsection{Optimization}\label{opt-KSANC}
	The optimization procedure of the proposed KSANC contains two stages. First, the teacher is trained from scratch by using labeled data. The attention-subnetwork with additional auxiliary layers ($i.e.,$ one fully-connected layer and a softmax output layer) and backbone-subnetwork are trained simultaneously by optimizing the two cross-entropy losses, while the auxiliary layers are removed during the process of transfer learning.
 Second, fixing the teacher network, the student and discriminator are updated under the framework of adversarial training, where the number of steps inside each component is simply set to $1$ in our experiments. Both student and discriminator are randomly initialized. We use Stochastic Gradient Descent (SGD) with momentum as the optimizer, and set the momentum as $0.9$, weight decay as ${\rm{1e-4}}$. The learning rate, initialized as ${\rm{1e-1}}$ and ${\rm{1e-3}}$ for student and discriminator separately, is multiplied by $0.1$ at three specific epoches during the training process. As for the regularization, having been examined the different values for weight factor $\mu$ in our experiments, we conclude that setting $\mu$ to $1$ is a good compromise for all evaluations empirically. For all experiments, we train on the standard training set and test on the validation set. Besides, data augmentation (random cropping and horizontal flipping) and normalization (subtracted and divided sequentially by mean and standard deviation of the training images) is applied to all the training images.
	
	%------------------------------------------------------------------------
	\section{Experiments}
	In this section, we start from describing the experimental settings, {\it i.e.,} datasets, evaluation measures, competitors and implementation details. Then ablation study is conducted to analyze the proposed method in detail. Finally, the performance comparison between the proposed method and state-of-the-art competitors on typical datasets are reported in subsection \ref{comp-sota}.
	
	\subsection{Experimental Setting}
	\textbf{Datasets:} We consider three image classification datasets: CIFAR-10, CIFAR-100, and ImageNet ILSVRC 2012. Both CIFAR-10 and CIFAR-100 contain $50$K training images and $10$K validation images, respectively. The ImageNet ILSVRC 2012 contains more than $1$ million training images from $1000$ object categories and $20$K validation images with each category including $20$ images. The image size of CIFARs and ImageNet is $32\times32$ and $224\times224$, separately.
	
	\textbf{Evaluation Measures:}
	To test the performance, we evaluate different models from the following two aspects: 1) the testing error of different student networks; 2) the convergence stability ($\boldsymbol{S}$) for training procedure. As for the former, the Top-1 error is calculated for all datasets, while the Top-5 error is additionally adopted for ImageNet. The testing error in ablation study is the average of twenty runs. The convergence stability is computed by the concussion range of the testing error:
	\begin{equation}
	\boldsymbol{S} = \text{Var}(\boldsymbol{\textit{\text{Err}}})
	\end{equation}
	where \text{Var} denotes the variance calculation, $\boldsymbol{\textit{\text{Err}}}$ is $[\textit{\text{err}}_{1}^{\max} - \textit{\text{err}}_{1}^{\min}, ... , \textit{\text{err}}_{E}^{\max} - \textit{\text{err}}_{E}^{\min}]$, $\textit{\text{err}}_{e}^{\max}$ and $\textit{\text{err}}_{e}^{\min}$ denote the maximum and the minimum error rate over twenty runs on $e$-th epoch ($e=1,\ldots,E$), respectively.

	\textbf{Competitors:}
	Since our proposed method is closely related with the attention transfer and the adversarial training, the following works should be included in our experiments. As for attention transfer, two representative knowledge transfer methods need to be analyzed: the FSP matrix \cite{agift} generated by the inner product of the features from two layers, and the attention map \cite{attmap} computed by the statistics of feature values across the channel dimension. These two approaches produce the transferred knowledge in a heuristic manner, while our model achieves a more compact one via the way of task-driven learning. For fair comparison, we adopt their representation of transferred knowledge into our framework.
	
	As for adversarial training, two recently representative methods, {\it i.e.,} adversarial network compression (ANC) \cite{ANC} and training student network with conditional adversarial networks (TSCAN) \cite{tstn}, are included. We implement the above two GAN based approaches on our own. Note that the backbone network in \cite{tstn}, {\it i.e.,} wide residual networks (WRN), is replaced by the ResNet, and the cross-entropy loss as well as the KD loss for student update are removed for fair comparison.

	Finally, for the purpose of comprehensive comparison, we introduce four additional knowledge distillation methods: Mimic learning with {\it{L}}2 loss (L2-Ba {\it{et al.}} \cite{ba2014deep}), distillation with soft targets via KL divergence (Hinton {\it{et al.}} \cite{hinton2015distilling}), knowledge transfer with FSP matrix (Yim {\it{et al.}} \cite{agift}), a deeper but thinner network (Fitnets, Romero {\it{et al.}} \cite{romero}). Four quantization methods: Weights binarization during training process except parameters update (Binary-Connect, Zhu, {\it{et al.}} \cite{courbariaux2015binaryconnect}), reducing the precision of the network weights to ternary values (Quantization, Courbariaux, {\it{et al.}} \cite{Zhu2016Trained}), binaryzation of the filters (BWN) and the additional input (XNOR) (Rastegari {\it{et al.}} \cite{rastegari2016xnor}).
	%Finally, Several compression strategies are introduced to compared with our comprehensive method on CIFAR-10, CIFAR-100 and ImageNet ILSVRC 2012. We choose six distillation \cite{Advances,hinton2015distilling,Romero2014FitNets,agift,ANC,tstn} and two quantization \cite{Courbariaux2015BinaryConnect,Zhu2016Trained} approaches for CIFAR-10. We examine the same six distillation methods for a comparison on CIFAR-100. We select two distillation \cite{Advances,ANC}, one quantization \cite{Rastegari2016XNOR} and one Compact Network Design \cite{Howard2017MobileNets} for a comparison on ImageNet ILSVRC 2012.
	
	\textbf{Implementation Details:}
	For CIFAR-10 and CIFAR-100, we set the pretrained teacher as ResNet-164, the student as ResNet-20\footnote{Note that the teacher and student networks do not restrict to one certain type, any other network, such as WRN, can be used in the same way.}, where preact block \cite{he2016identity} is employed since it is currently the standard architecture for recognition. We select the minibatch size as $64$ and total train epoch as $600$ with the learning rate multiplied by $0.1$ at epoch $240$ and epoch $480$. We adjust the teacher network to ResNet-152 like ANC \cite{ANC} and TSCAN \cite{tstn} since the pretrain model is available, the student network is changed to ResNet-50/18, the mini-batch size is set to $128$, and the total epoch is altered to $120$ for ImageNet dataset, where the learning rate is divided by $10$ at epoch $30$, $60$ and $90$. Our implementation is based on Pytorch, with $1$ and $4$ NVIDIA GTX 1080ti GPU for CIFAR-10/100 and ImageNet, separately.
	
	\subsection{Ablation Study}
	In this subsection, we compare the different strategies of attention transfer, then analyse the impact of different combinations of loss function for the discriminator. Finally, the advantages of the intermediate supervision are discussed.
	
	\textbf{Comparison of Attention Transfer:}
	We aim to demonstrate that the task-driven attention mechanism is a more effective way to squeeze the knowledge transferred from teacher to student than the FSP matrices in \cite{agift} and the attention map in \cite{attmap}. To do so, we match these three types of representation between teacher and student network by using $L2$ distance as the intermediate supervision, and then observe the classification error on the two benchmark datasets.
	
	As illustrated in Table \ref{tab:diffIS}, it demonstrates that the backbone loss $L_{b}$, combining with intermediate loss $L_{is}$ with any form of transferred knowledge (appointed in bracket in the first column in Table \ref{tab:diffIS}), could facilitate the training of student network, especially our squeezed transferred knowledge, which outperforms other representations by a significant margin. It indicates that, by integrating the task-driven attention scheme, our squeezed knowledge is more suitable to be adapted into the small scale network.
	
		\begin{table}[!htb]
		\begin{center}
			\begin{tabular}{|p{3.5cm}|p{1.5cm}<{\centering}|p{1.7cm}<{\centering}|}
				\hline
				\multirow{2}*{Loss composition} & \multicolumn{2}{|c|}{Error $[\%]$} \\
				\cline{2-3}
				~ & CIFAR-10 & CIFAR-100 \\
				\hline
				${L_b}$ & $8.19$ & $32.60$\\
				\hline
				${L_b + L_{is}}$ \cite{attmap} & $ 7.97 $ & $ 32.31 $\\
				\hline
				${L_b + L_{is}}$ \cite{agift} & $ 8.03 $ & $ 32.45 $\\
				\hline
				${L_b+L_{is}}$[ours]&$ 7.55 $&$ 31.97 $\\
				\hline
			\end{tabular}
		\end{center}
		\caption{The evaluation of different attention transfer methods. Same backbone loss $L_b$ is applied. The effect of different intermediate losses $L_{is}$ \cite{attmap}, $L_{is}$ \cite{agift}, and our proposed $L_{is}$ is studied. }
		\label{tab:diffIS}
	\end{table}

	\textbf{Loss Functions for Discriminator: } Since the proposed KSANC is built upon the adversarial training framework, recall the subsection \ref{loss-KSANC}, it is necessary for us to find the reasonable combination of loss functions for the discriminator. As it is shown in Table \ref{tab:diffadvloss}, generally speaking, any adversarial loss can improve the performance of student network. Specifically, by comparing the difference between the line $1$ and $2$, and the difference between the line $1$ and $3$, either category regularizer $L_{adv}^{C}$ or discriminator regularizer $L_{reg}$ could boost the performance slightly. However, their joint constraint (see the last row in Table \ref{tab:diffadvloss}) will lead to a remarkable improvement, which indicates that both $L_{adv}^{C}$ and $L_{reg}$ play critical roles in our adversarial training model.

	\begin{table}[!htb]
		\begin{center}
			\begin{tabular}{|p{3.5cm}|p{1.5cm}<{\centering}|p{1.7cm}<{\centering}|}
				\hline
				\multirow{2}*{Loss composition} & \multicolumn{2}{|c|}{Error $[\%]$} \\
				\cline{2-3}
				~ & CIFAR-10 & CIFAR-100 \\
				\hline
				${L_b+L_{is}}$&$7.55$&$31.97$\\
				\hline
				${L_b + L_{is} + L_{adv}^o + L_{reg}}$ & $7.47$ & $31.62$\\
				\hline
				${L_b + L_{is} + L_{adv}^o + L_{adv}^{C}}$ & $7.39$ & $31.71$\\
				\hline
				${L_b+L_{is}+L_{adv}}$&$7.32$&$31.42$\\
				\hline
			\end{tabular}
		\end{center}
		\caption{The evaluation of different components of adversarial loss. Same backbone loss $L_b$ and intermediate loss $L_{is}$ are employed. The effect of different advisarial loss components $L_{adv}^o$, $L_{reg}$ and $L_{adv}^{C}$ is evaluated.}
		\label{tab:diffadvloss}
	\end{table}
	
	\textbf{Benefits of Intermediate Supervision: } We look into the effect of enabling and disabling different loss components of KSANC model, as shown in Table \ref{tab:loss}. We can see that even merely using backbone loss ${L_b}$ could be able to obtain a better effect than student network without any knowledge transfer (directly supervised learning by using sample-label pair). Moreover, both ${L_{adv}}$ and ${L_{is}}$ could improve the performance of student network.

Interestingly, utilizing ${L_{is}}$ get better result than ${L_{adv}}$. We give the explanation as follows: Recall the subsection \ref{loss-KSANC}, both loss functions $L_b$ and $L_{adv}$ give service to the result-oriented learning, which is naturally without the guidance from intermediate supervision (see line $4$ in Table \ref{tab:loss}). As a result, significant improvement can not be acquired by incorporating another result-oriented loss function. On the contrary, further improvement can be realized by adding the intermediate loss $L_{is}$ to $L_b$ (see line $3$ in Table \ref{tab:loss}), which is a absolutely evidence that both of the them can complement each other. Furthermore, the final approach combining all of the loss components preforms the best, this is attributed to the advantage of the adversarial training.
	
	\begin{table}[!htbp]
		\begin{center}
			\begin{tabular}{|p{3.9cm}|p{1.5cm}<{\centering}|p{1.7cm}<{\centering}|}
				\hline
				\multirow{2}*{Loss composition} & \multicolumn{2}{|c|}{Error $[\%]$} \\
				\cline{2-3}
				~ & CIFAR-10 & CIFAR-100 \\
				%Loss composition&CIFAR-10&CIFAR-100\\
				\hline
				supervised learning & $ 8.58 $ & $ 33.36 $\\
				\hline
				${L_b}$&$ 8.19 $&$ 32.60 $\\
				\hline
				${L_b+L_{is}}$&$ 7.55 $&$ 31.97 $\\
				\hline
				${L_b+L_{adv}}$&$ 7.72 $&$ 32.17 $\\
				\hline
				${L_b+L_{adv}+L_{is}}$&$ 7.32 $&$ 31.42 $\\
				\hline
			\end{tabular}
		\end{center}
		\caption{The effect of different components of the loss in KSANC.}
		\label{tab:loss}
	\end{table}
	
\begin{figure}[htbp]
		\centering
		\begin{minipage}[t]{0.48\textwidth}
			\centering
			\includegraphics[width=9cm]{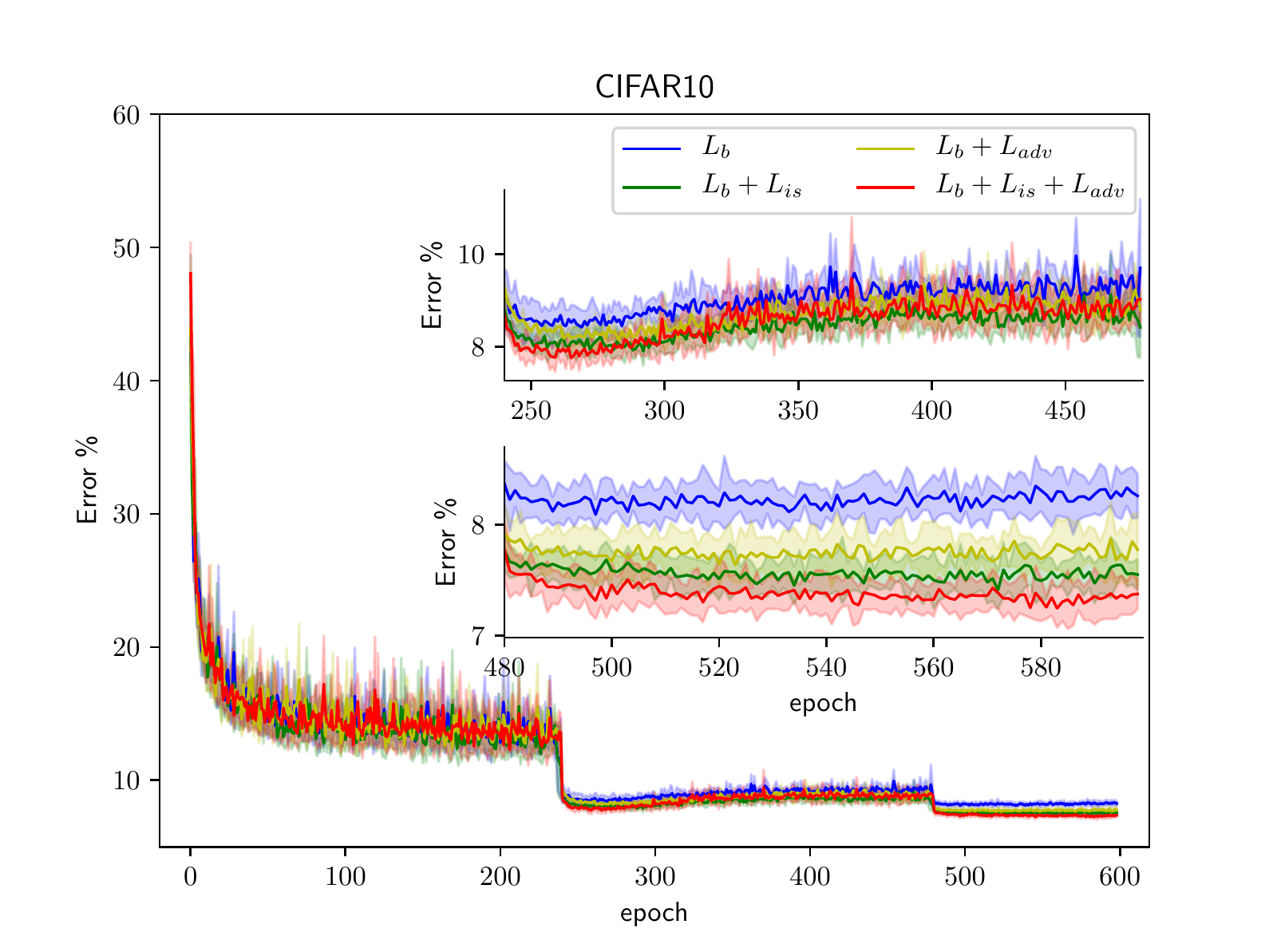}
			\caption{Training of four different models on CIFAR-10, where X axis denotes training epochs and Y axis denotes testing error.}
			\label{fig:stb-cifar10}
		\end{minipage}
	\end{figure}

\begin{figure}[htbp]
		\begin{minipage}[t]{0.48\textwidth}
			\centering
			\includegraphics[width=9cm]{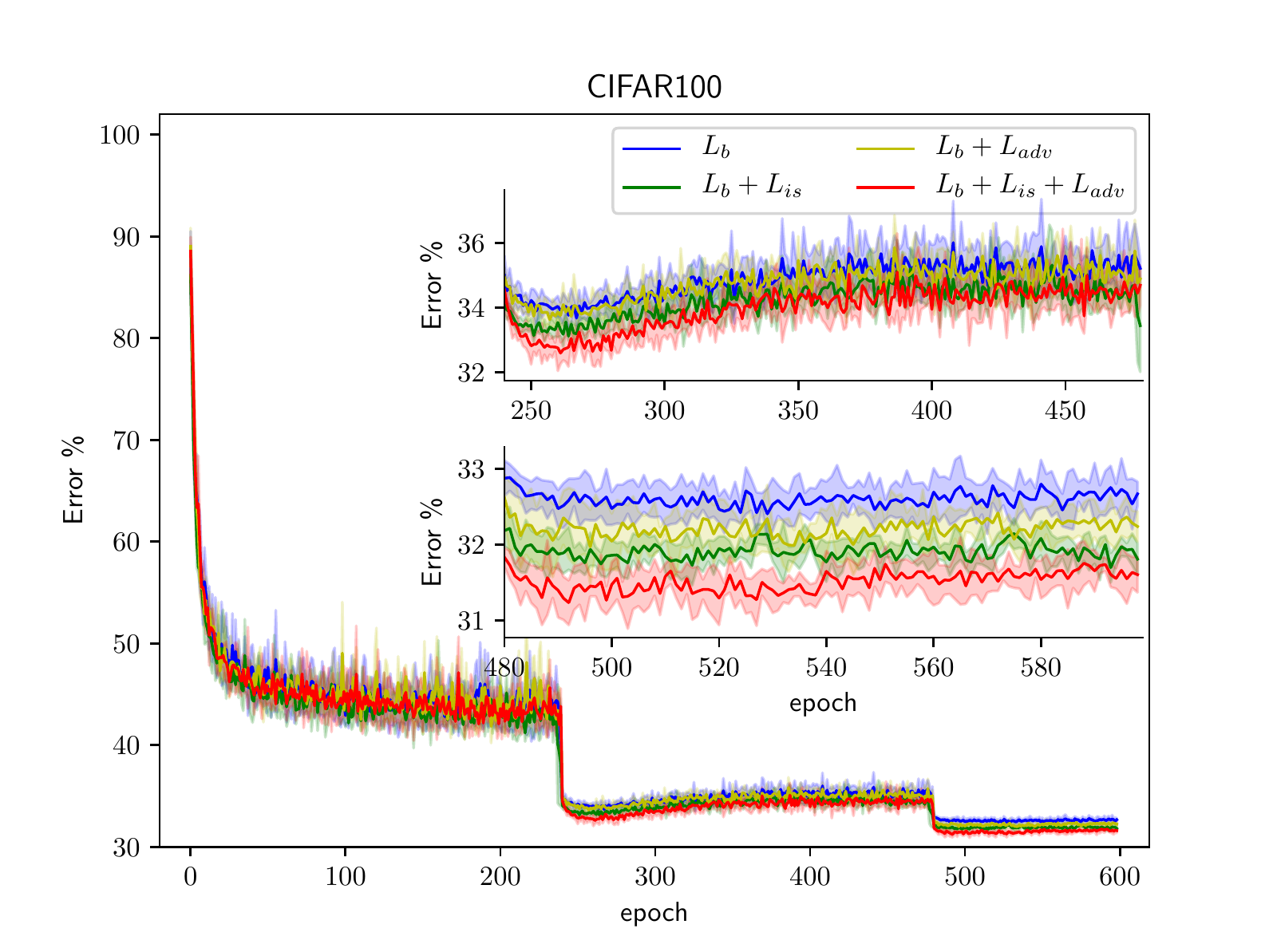}
			\caption{Training of four different models on CIFAR-100, where X axis denotes training epochs and Y axis denotes testing error.}
			\label{fig:stb-cifar100}
		\end{minipage}
	\end{figure}

	Intuitively, the benefit of combining the result-oriented and process-oriented learning will be presented in the training procedure, {\it e.g.,} the training curve. Fig. \ref{fig:stb-cifar10} and Fig. \ref{fig:stb-cifar100} represents the test error of different models over time on CIFAR-10 and CIFAR-100, respectively. Our model ($L_b+L_{adv}+L_{is}$, the red line in Fig. \ref{fig:stb-cifar10} and Fig. \ref{fig:stb-cifar100}) has a relatively lower testing error during the training process, especially after epoch $240$, see the zoom-in windows in the upper right.
	
	Moreover, we select the last $100$ epochs to calculate the convergence stability of different models, which is illustrated in Table \ref{tab:var}. By comparing the first two lines and last two lines respectively, the convergence stability becomes more apparent after integrating the loss function of intermediate supervision, which further indicates that the process-oriented learning could improve the stability of transfer learning.

	\begin{comment}
	\begin{figure}[htbp]
	%\centering %?��?D
	\subfigure[CIFAR-10]{%�̨���???������?
	\begin{minipage}{4.1cm}
	%\centering%������??��?D
	\includegraphics[scale=0.286]{./fig/fig_4.pdf}%��?pic.jpg��?0.5��?�䨮D?��?3?
	\end{minipage}}
	\subfigure[CIFAR-100]{%�̨�?t??������?
	\begin{minipage}{4.1cm}
	%\centering %������??��?D
	\includegraphics[scale=0.286]{./fig/fig_5.pdf} %��?pic.jpg��?0.5��?�䨮D?��?3?
	\end{minipage}}
	\caption{The training curevs on CIFAR-10 and CIFAR-100. We show the testing error rate of different models.} %
	\label{fig:stb}%��???��y��?����??
	\end{figure}
	\end{comment}


	\begin{table}[!htb]
		\begin{center}
			\begin{tabular}{|p{3.5cm}|p{1.5cm}<{\centering}|p{1.7cm}<{\centering}|}
				\hline
				\multirow{2}*{Model} & \multicolumn{2}{|c|}{Stability} \\
				\cline{2-3}
				~ & CIFAR-10 & CIFAR-100 \\
				\hline
				${L_b}$ & ${\rm{2.46e-3}}$ & ${\rm{5.12e-3}}$\\
				\hline
				${L_b+L_{is}}$ & ${\rm{2.13e-3}}$ & ${\rm{4.42e-3}}$\\
				\hline
				${L_b+L_{adv}}$ & ${\rm{3.02e-3}}$ & ${\rm{6.41e-3}}$\\
				\hline
				${L_b+L_{adv}+L_{is}}$ & ${\rm{2.34e-3}}$ & ${\rm{5.67e-3}}$\\
				\hline 	
			\end{tabular}
		\end{center}
		\caption{ Convergence Stability. The variance of testing error concussion range on CIFAR-10 and CIFAR-100 through last $100$ epochs are shown. Different loss components for KSANC are used.}
		\label{tab:var}
	\end{table}
	
	\subsection{Comparison with State-of-the-art} \label{comp-sota}
	
	We first compare our model with several cutting edge compression approaches, including six distillation algorithms and four quantization ones. Eight of them, {\it i.e.,} six distillations and two quantization approaches, are available for CIFAR-10 and CIFAR-100 datasets, other unavailable methods, whose results are not provided by their authors, are omitted in the reported table. Similar way is adopted in the experiment for ImageNet dataset.
	
	\begin{table}[!ht]
		\begin{center}
			\begin{tabular}{|p{3.2cm}|p{1.0cm}<{\centering}|p{1cm}<{\centering}|p{1cm}<{\centering}|}
				\hline
				\multirow{2}*{Model} & \multirow{2}*{Param} & \multicolumn{2}{|c|}{Error $[\%]$}  \\
				\cline{3-4}
				~ & ~ & CIFAR-10 & CIFAR-100 \\
				\hline
				Supervised teacher RN-164 & $2.6$M & $6.57$ & $27.76$ \\
				Supervised student RN-20 & $0.27$M & $8.58$ & $33.36$ \\
				\hline
				Yim \cite{agift} & $0.27$M & $11.30$ & $36.67$\\
				L2-Ba \cite{ba2014deep} & $0.27$M & $9.07$ & $32.79$\\
				%\hline
				Hinton \cite{hinton2015distilling} & $0.27$M & $8.88$ & $33.34$\\
				%\hline
				FitNets \cite{romero} & $2.5$M & $8.39$ & $35.04$\\
				%\hline
				%Yim $et.al$\cite{szegedy2016rethinking} & $0.27$M & $11.30$ & $36.67$\\
				%\hline
				Quantization \cite{Zhu2016Trained} & $0.27$M & $8.87$ & $-$\\  %
				%\hline
				Binary-Connect \cite{courbariaux2015binaryconnect} & $15.20$M & $8.27$ & $-$\\
				%\hline
				ANC \cite{ANC} & $0.27$M & $8.08$ & $32.45$\\
				%\hline
				TSCAN \cite{tstn} & $0.27$M & ${7.93}$ & ${32.57}$\\
				\hline
				KSANC & $0.27$M & ${\bold{7.32}}$ & ${\bold{31.42}}$\\
				\hline
			\end{tabular}
		\end{center}
		\caption{Comparison with state-of-the-art methods on CIFAR-10 and CIFAR-100. Teacher and student network are RN-164 and RN-20 respectively. (RN denotes ResNet.)}
		\label{tab:sofa-cifar}
	\end{table}

	\begin{table}[!ht] %
		\begin{center}
			\begin{tabular}{|p{3.2cm}|p{1.cm}<{\centering}|p{1cm}<{\centering}|p{1cm}<{\centering}|}
				\hline
				\multirow{2}*{Model} & \multirow{2}*{Param} & \multicolumn{2}{|c|}{Error $[\%]$}  \\
				\cline{3-4}
				~ & ~ & Top-1 & Top-5 \\
				\hline
				Supervised teacher RN-152 & $58.21$M & $27.63$ & $5.90$ \\
				Supervised student RN-50 & $37.49$M & $30.30$ & $10.61$ \\
				Supervised student RN-18 & $13.95$M & $43.33$ & $20.11$ \\
				\hline
				XNOR \cite{rastegari2016xnor} (RN-18) & $13.95$M & $48.80$ & $26.80$ \\
				%\hline
				BWN \cite{rastegari2016xnor} (RN-18) & $13.95$M & $39.20$ & $17.00$ \\
				%\hline
				L2-Ba \cite{ba2014deep} (RN-18) & $13.95$M & $33.28$ & $11.86$\\
				ANC \cite{ANC} (RN-18) & $13.95$M & $32.89$ & $11.72$\\
				TSCAN \cite{tstn} (RN-18) & $37.49$M & ${32.72}$ & ${11.49}$\\
				KSANC(RN-18) & $13.95$M & ${\bold{31.47}}$ & ${\bold{10.93}}$ \\
				\hline
				L2-Ba \cite{ba2014deep} (RN-50) & $37.49$M & $27.99$ & $9.46$\\
				%\hline
				ANC \cite{ANC} (RN-50) & $37.49$M & $27.48$ & $8.75$\\
				%\hline
				TSCAN \cite{tstn} (RN-50) & $37.49$M & ${27.39}$ & ${8.53}$\\
				%\hline
				KSANC (RN-50) & $37.49$M & ${\bold{26.79}}$ & ${\bold{8.01}}$ \\
				\hline
			\end{tabular}
		\end{center}
		\caption{Comparison with state-of-the-art methods on ImageNet. Teacher network is RN-152. (RN denotes ResNet.)}
		\label{tab:sofa-imagenet}
	\end{table}

	As shown in the Table \ref{tab:sofa-cifar}, the deep teacher preforms much better than the shallow student network with supervised learning (line $2$ in Table \ref{tab:sofa-cifar}), and the error rate of small network learned by using distillation models is bounded by the teacher's performance, as expected. Both distillation and quantization approaches obtain relatively good performance with a small model size. Specifically, two GAN based competitors (ANC \cite{ANC} and TSCAN \cite{tstn}) achieve desirable results which outperform the supervised learning for student with $0.5\%$ and $0.65\%$, respectively. Obviously, the proposed KSANC further boost the capability of student network. More noticeable improvements can be seen on the CIFAR-100 dataset. To sum up, we can see that our method acquires the lowest error with the same or less number of parameters, which demonstrates that our model benefits from effective representation of transferred knowledge and intermediate supervision.

More adequate evidence is provided by the comparison on large-scale dataset, {\it i.e.,} ImageNet. Two widely used networks, ResNet-50 and ResNet-18, are employed as the student network, and the comparison results are presented in Table \ref{tab:sofa-imagenet}. By analyzing the results in the first group (line $1$ to $3$), we can deduce that ResNet-152 might contain redundancy since the difference between ResNet-152 and ResNet-50 is only $2.67\%$. In that case, the ANC, TSCAN, and our method could obtain desirable results which even beat the teacher network, as the ResNet-50 is a more concise architecture trained by the three distillation methods. 

When the size of student network becomes smaller (ResNet-18), the error of the proposed method increases nearly $4.7\%$ (from $26.73\%$ to $31.47\%$), where our assumption that, ``the small network can not perfectly mimic a large on especially when there exists significant difference in the number of layer.'', can be confirmed. Nevertheless, the proposed method obtains the performance gain w.r.t the second best (TSCAN, see the last two lines in the second group of Table \ref{tab:sofa-imagenet}) better than that in the case of ResNet-50. This indicates that, the transferred knowledge in our model is much more suitable for injecting into the small scale network.

	\section{Conclusion}
	In this paper, a novel knowledge transfer method is proposed via the framework of knowledge squeezed adversarial training for distillation. To inherit the information from teacher to student effectively and compactly, the task-driven attention mechanism is designed to squeeze the knowledge for intermediate supervision. Moreover, the category regularizer and discriminator regularizer are introduced to improve the adversarial training. Extensive evaluation of our KSANC is conducted on three challenging image classification datasets, where a clear outperformance over contemporary state-of-the-art methods is achieved. Additionally, The experimental results demonstrate that, the proposed attention transfer method can squeeze the intermediate information in a more compact form, which further facilitate the convergence stability via the intermediate supervision.

	{\small
		\bibliographystyle{ieee}
		\bibliography{egbib}
	}
	
\end{document}